\documentclass[10pt,twocolumn,letterpaper]{article}

\usepackage{cvpr}              
\usepackage{pifont}

\usepackage[dvipsnames]{xcolor}

\newcommand{\algname}{PostoMETRO\xspace}

\definecolor{cvprblue}{rgb}{0.21,0.49,0.74}
\usepackage[pagebackref,breaklinks,colorlinks,citecolor=cvprblue]{hyperref}

\title{PostoMETRO: Pose Token Enhanced Mesh Transformer for Robust 3D Human Mesh Recovery}

\author{
Wendi Yang$^{1,2,}$\thanks{These authors contributed equally to this work} \qquad 
Zihang Jiang$^{1,2,}$\footnotemark[1]\qquad Shang Zhao$^{1,2}$\qquad
S. Kevin Zhou$^{1,2,}$\thanks{Corresponding author}\\
\textsuperscript{1}School of Biomedical Engineering \& Suzhou Institute for Advanced Research,\\
University of Science and Technology of China\\
\textsuperscript{2}Center for Medical Imaging, Robotics, Analytic Computing \& Learning (MIRACLE),\\
Suzhou Institute for Advanced Research, University of Science and Technology of China\\
{\tt\small yangwendi@mail.ustc.edu.cn} \\
{\tt\small \{jzh0103, shangzhao, skevinzhou\}@ustc.edu.cn}}

\begin{document}
\nolinenumbers
\maketitle
\begin{abstract}
\vspace{-4mm}
With the recent advancements in single-image-based human mesh recovery, there is a growing interest in enhancing its performance in certain extreme scenarios, such as occlusion, while maintaining overall model accuracy.
Although obtaining accurately annotated 3D human poses under occlusion is challenging, there is still a wealth of rich and precise 2D pose annotations that can be leveraged.
However, existing works mostly focus on directly leveraging 2D pose coordinates to estimate 3D pose and mesh. 
In this paper, we present \algname (\textbf{Pos}e \textbf{to}ken enhanced \textbf{ME}sh \textbf{TR}ansf\textbf{O}rmer), which integrates occlusion-resilient 2D pose representation into transformers in a token-wise manner. 
Utilizing a specialized pose tokenizer, we efficiently condense 2D pose data to a compact sequence of pose tokens and feed them to the transformer together with the image tokens. 
This process not only ensures a rich depiction of texture from the image but also fosters a robust integration of pose and image information. 
Subsequently, these combined tokens are queried by vertex and joint tokens to decode 3D coordinates of mesh vertices and human joints. 
Facilitated by the robust pose token representation and the effective combination, we are able to produce more precise 3D coordinates, even under extreme scenarios like occlusion. 
Experiments on both standard and occlusion-specific benchmarks demonstrate the effectiveness of \algname. 
Qualitative results further illustrate the clarity of how 2D pose can help 3D reconstruction. 
Code will be made available.
\vspace{-8mm}
\end{abstract}    
\section{Introduction}
\label{sec:intro}
The 3D human pose and shape estimation (3DHPSE) task has been increasingly valued in the field of computer vision, due to the growing number of downstream tasks, such as virtual humans, VR, and AR. 
Among these, due to the inherent challenges of camera calibration in outdoor scenes, the 3DHPSE task under a monocular camera has received extensive research attention. 
However, accompanying this are the challenging issues specific to monocular scenes, such as depth ambiguity and occlusion problems.

\begin{figure}[t]
  \centering
   \includegraphics[width=1.0\linewidth]{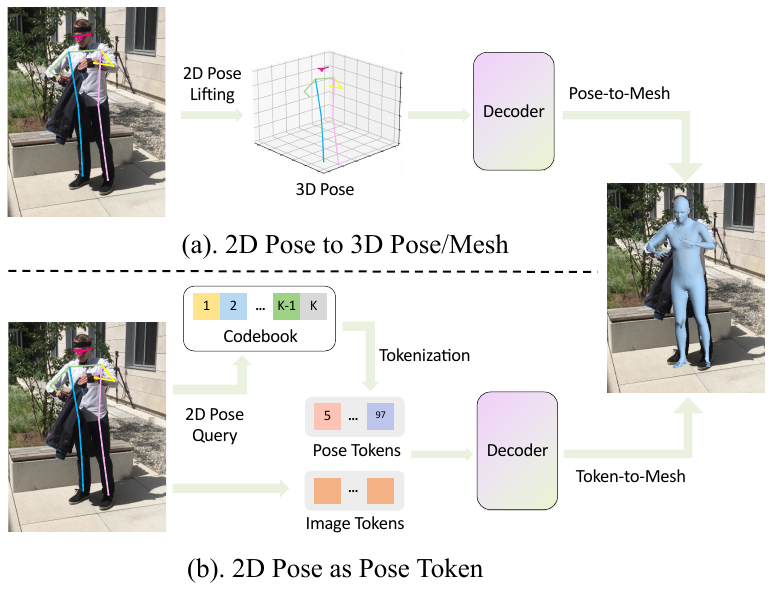}
   \caption{(a) Traditional approach of utilizing 2D pose information by converting it into 3D pose and subsequently generating a 3D mesh. (b) Transforming 2D pose into a sequence of tokens and integrating it with image tokens to generate a 3D mesh.}
   \vspace{-4mm}
   \label{fig:pose-paradigm}
\end{figure}

One way of handling 3DHPSE task is to glean 2D clues on images, such as 2D human pose~\cite{choi2020pose2mesh} and mesh-aligned 2D human texture~\cite{zhang2021pymaf,kim2023PointHMR}, to reconstruct 3D human mesh by advanced neural networks~\cite{kipf2016gcn, dosovitskiy2020vit}. 
Despite the high accuracy these paradigms demonstrate on some indoor datasets~\cite{ionescu2013human3}, their performance drastically drops in the presence of severe occlusion. This scenario spoils the alignment between human mesh vertices and image pixels, breaking the pixel-wise dependency needed for decoding 3D coordinates on the human mesh.
To mitigate the impact of such image-mesh misalignment issue caused by occlusion, many works~\cite{kanazawa2018hmr,kocabas2021pare,kolotouros2019spin,choi2022_3DCrowdNet,choi2020pose2mesh} propose to incorporate occlusion-robust prior knowledge in order to enhance model performance. 
A common practice to counteract the negative effects of occlusion is to utilize the parameterized model SMPL~\cite{SMPL:2015}, since the SMPL model encodes human pose and shape priors into compact features and hence can produce plausible results when occlusion happens. 
Although lots of works~\cite{kocabas2021pare,kolotouros2019spin,kanazawa2018hmr} attempt to recover human mesh by regressing SMPL parameters, incorporating SMPL prior knowledge from its feature space into a non-parametric model is still a research challenge. 
The common approach of existing SOTA methods~\cite{choi2020pose2mesh, choi2022_3DCrowdNet} is to utilize 2D poses with a focus on incorporating occlusion-robust clues like 2D human pose. 
This is done by obtaining an intermediate 3D human representation, \textit{i.e.}, 3D pose, which is then utilized for robust 3D human mesh estimation as shown in  \cref{fig:pose-paradigm} (a). 
Unlike SMPL-based methods, they do not store human pose prior knowledge in a high-dimensional feature space but in 2D coordinates. 
Although such hand-crafted 2D-to-3D methods achieve good controllability and demonstrate comparable performances, these processes still face challenges. 
For instance, inherent uncertainty, \textit{i.e.}, depth ambiguity in \textit{z}-axis and inaccuracy in \textit{x,y}-axis, exists in obtaining the 3D pose for constructing the 3D mesh, which may lead to significant errors when recovering the human mesh from such intermediate representations.

Inspired by aforementioned ideas, in this paper, we present \algname, a novel paradigm to improve the performance of non-parametric model under occlusion scenarios, which is done by encoding occlusion-robust prior knowledge into the model in a cost-effective way. 
We observe that the explicit utilization of human 2D pose to get 3D mesh suffers from erroneous intermediate 3D pose. 
In light of recent progress in compressing 2D human body poses down to structural representation~\cite{geng2023pct} and the remarkable performance of the models leveraging the SMPL model as a prior knowledge source, we directly compress 2D human pose into compact token-wise feature by using a VQ-VAE, represented in \cref{fig:pose-paradigm} (b). 
It is worth noting that during this process, a 2D pose is compressed into another representation in a data-driven manner and no manual design is required. 
Such transformation enables the compression of 2D poses into the feature space for end-to-end utilization, without going through any erroneous intermediate representation. 
Combining pose tokens with image tokens encoded by CNNs~\cite{he2016resnet, wang2020hrnet}, we then use \textbf{a transformer-based encoder} to further facilitate the interaction between fine-grained image and pose tokens. 
To avoid the model being misled by inaccurate 2D poses, we further incorporate the uncertainty from the 2D poses as part of pose tokens. 
To decode a human mesh from image and pose tokens, following the prior practice~\cite{lin2021Metro,cho2022FastMetro}, we pre-define human mesh vertices and joints as tokens, \textbf{a transformer-based decoder} is used to accomplish interaction among these predefined tokens, image tokens, and pose tokens. 
A lightweight linear layer is used to transform point information from feature level to 3D coordinates. The overall pipeline is in ~\cref{fig:model-arch}.

We benchmark our model on 3DPW~\cite{von2018pw3d}, 3DPW-OCC~\cite{zhang2020pw3docc}, 3DPW-PC~\cite{sun2021romp} and 3DOH~\cite{zhang2020pw3docc}. 
Our experiments show that our model yields state-of-the-art results, proving the proficiency and adaptability of our model on datasets involving object-occlusion, person-occlusion, and non-occlusion scenarios. 
Finally, we offer ablation studies and qualitative results, which demonstrate the effectiveness of our occlusion-robust 2D pose integration strategy. 
Our main contributions are summarized as follows:
\begin{itemize}
    \item We propose \algname, a novel framework to incorporate 2D pose into transformers to help 3D human mesh estimation, which is done by transforming 2D pose as pose tokens.
    \item Our method demonstrates occlusion-robustness and strong pixel-aligned accuracy on both standard benchmarks and occlusion-specific benchmarks.
    \item Our approach can generalize to different occlusion scenarios including object-occlusion and person-occlusion without any dedicated design.
\end{itemize}
\section{Related Work}
\label{sec:related_work}
\subsection{3D Human Pose and Shape Estimation}
To estimate 3D human body, early works~\cite{bogo2016smplify, lassner2017up3d, pavlakos2019smplx, varol2018bodynet} adopt optimization and use 2D keypoints and human silhouette to output parameters of parametric models~\cite{SMPL:2015, pavlakos2019smplx, anguelov2005scape, xu2020ghum}. 
SMPLify~\cite{bogo2016smplify} was the first method to fit the SMPL~\cite{SMPL:2015} model to the given 2D keypoints. 
These methods are slow and not robust as the optimization procedure is usually time-consuming and sensitive to initialization.
Recently, learning-based methods~\cite{kanazawa2018hmr,kolotouros2019spin,zhang2021pymaf,kocabas2021pare,lin2021Metro,cho2022FastMetro,choi2020pose2mesh} have gained wide attention since they can reconstruct human model parameters directly from a single image.
~\cite{kanazawa2018hmr, kolotouros2019spin, zhang2021pymaf, kocabas2021pare} propose to use image features to directly regress the SMPL parameters. 
To facilitate training, extra supervisory signals are used, \textit{e.g.}, 2D keypoints~\cite{kanazawa2018hmr} and human body segmentation map~\cite{kocabas2021pare}. 
Kolotouros \textit{et al.}~\cite{kolotouros2019spin} combine 2D keypoints and SMPLify model to achieve extra supervision.

While model-based approaches have made remarkable progress in 3D human mesh reconstruction, their performances are constrained by pre-defined body models, making it challenging to extend them to other applications. 
To tackle this, some works~\cite{choi2020pose2mesh,lin2021Metro,cho2022FastMetro} proposed to use advanced neural network architecture~\cite{kipf2016gcn,dosovitskiy2020vit} to handle 3DHPSE problem without relying on SMPL~\cite{SMPL:2015}. 
Choi \textit{et al.}~\cite{choi2020pose2mesh} utilize GCN~\cite{kipf2016gcn} to regress human mesh by predicting 3D coordinates of mesh vertices. 
Lin \textit{et al.}~\cite{lin2021Metro} leverage strong capability of vision transformer encoder~\cite{dosovitskiy2020vit} to model long-range dependency among human mesh vertices. 
Cho \textit{et al.}~\cite{cho2022FastMetro} enhance the transformer architecture by transitioning from an encoder-only to an encoder-decoder framework. 

\vspace{-1mm}
\subsection{Occlusion Handling}
\vspace{-1mm}
Solving the occlusion problem on single image can be very challenging. 
This is because one cannot utilize clues from other views to disambiguate the uncertainty of target object, and the occluders can vary from random object to real person. 
One straightforward way to handle occlusion is to rely on data augmentation during training to synthesize occlusion~\cite{kocabas2021pare,li2022cliff}. 
However, due to the texture difference between synthetic data and real occluders, such methods can not fully guarantee the performance of model in the real world. 
Zhang \textit{et al.}~\cite{zhang2020pw3docc} propose to cast an occlusion disambiguation problem into an image-inpainting problem using UV maps, but their work needs high-quality saliency maps and UV maps. 
Kocabas \textit{et al.}~\cite{kocabas2021pare} propose to infer unseen parts from visible parts, but this approach also relies on the model's capability to capture the high-quality visible parts. 
Similar to our work, Choi \textit{et al.}~\cite{choi2022_3DCrowdNet} propose to leverage 2D human pose to mitigate the influence of occlusion. 
However, their approach only focuses on solving person-occlusion while our approach is suitable for object-occlusion and person-occlusion at the same time.

\vspace{-1mm}
\subsection{Structral Pose Representation}
\vspace{-1mm}
Leveraging 2D pose in 3DHPSE task has always been mainstream, but most works~\cite{kanazawa2018hmr,kolotouros2019spin,joo2021exemplar, moon2022neuralannot} either just use 2D pose as extra supervisory signal or to estimate accurate SMPL parameters to help training. 
Choi \textit{et al.}~\cite{choi2022_3DCrowdNet} adopt 2D pose to calibrate CNN and extract fine-grained feature to improve model performance under person-occlusion scenarios. 
However, due to the uncertain 3D pose used to get 3D mesh, the estimated human body can be erroneous. 
Recently, transforming 2D poses into sequences of tokens has become popular due to the powerful token-level modeling ability of the transformer. 
Lin \textit{et al.}~\cite{lin2022mpt} leverage large-scale MoCap dataset~\cite{mahmood2019amass} to generate 2D pose heatmaps and then pretrain transformers to recover human mesh by 2D pose token transformed from heatmaps. 
Their approach, \textit{i.e.}, MPT, neglects the inherent 2D-to-3D ambiguity since they only use 2D pose token as input, while our approach uses image token to compensate 2D-to-3D ambiguity. Besides that, the scale of our pretraining datasets is more lightweight than that of MPT.
Qiu \textit{et al.}~\cite{qiu2023psvt} combine token-level pose representation with image token to refine SMPL parameters. 
However, their approach aims for multi-person scenarios so that the pose token might contain pose information of multiple individuals, which brings complexities, while our approach does not have such problems since we focus on single-person scenarios. 
Recently, VQ-VAE~\cite{van2017vqvae} has been widely used for explicitly encoding pose prior in learnable embedding~\cite{zhang2023T2M-GPT, guo2022tm2t, gong2023tm2d, geng2023pct}. 
However, most of them~\cite{zhang2023T2M-GPT, guo2022tm2t, gong2023tm2d} focus on using such representation for motion generation, which may not demand the same level of precision as discrimination tasks. 
Closely related to our work, Geng \textit{et al.}~\cite{geng2023pct} propose to represent 2D poses using entries from VQ-VAE. 
However, their study is confined to 2D human pose regression, constrained by the substantial gap between 2D and 3D. 
Consequently, their approach cannot be directly applied to 3DHPSE tasks.
\section{Method}
\begin{figure*}[t]
  \centering
   \includegraphics[width=1\linewidth]{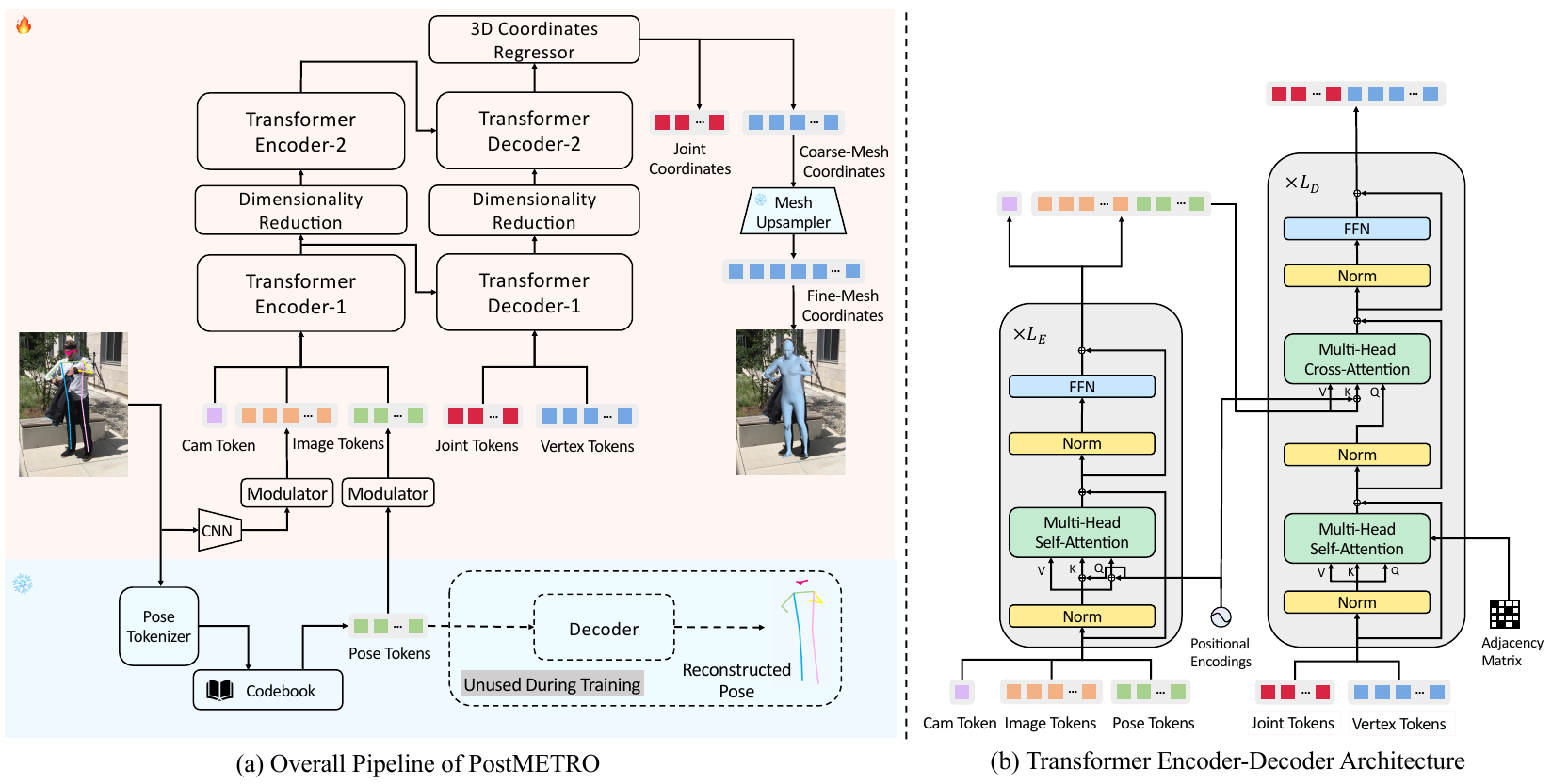}
   \caption{\textbf{Overall pipeline of \algname (a) and architecture of transformer encoder-decoder (b).} Before training transformers, we learn a ``pose tokenizer" in VQ-VAE fashion, where a human image will be tokenized to discrete pose tokens, which can be reconstructed to a 2D pose by a decoder, according to the learned codebook. During training transformers, we freeze pose tokenizer and mesh upsampler and train transformers by concatenating pose token to camera token and image tokens as input for transformer encoders. For transformer decoders, learnable joint tokens and vertex tokens are fed as input. Following~\cite{cho2022FastMetro, kim2023PointHMR}, dimensionality reduction and adjacency matrix are used for effectiveness. Note that pose confidence for 2D pose is omitted for simplicity.}
   \vspace{-4mm}
   \label{fig:model-arch}
\end{figure*}

\subsection{Preliminary: Pose Tokenizer}
\label{sec:pose_token}
Following the progress in the field of natural language processing~\cite{devlin2018bert} and image processing~\cite{ramesh2021dalle}, we transform 2D pose into discrete token sequence by learning a "pose tokenizer". 
More specifically, a cropped human image $X \in \mathbb{R}^{C\times H \times W}$ is fed into the pose tokenizer, resulting in a sequence of human pose tokens ${Z}=[z_{1},\ldots,z_{N}]^{T} \in \mathbb{R}^{N\times d_{p}}$ corresponding to the target human, where each token feature $z_{i}$ is extracted from codebook $\mathbf{C} \in \mathbb{R}^{V \times d_{p}} = [c_{1},...,c_{V}]^{T}$.

We learn the pose tokenizer using a vector quantised-variational autoencoder (VQ-VAE) following~\cite{geng2023pct}. 
To facilitate training, a two-stage training scheme is adopted. In stage one, an MLP-Mixer~\cite{tolstikhin2021mlpmixer} based \textit{pose encoder} is employed to compress 2D coordinates $P \in \mathbb{R}^{J \times 2}$ into a discrete token sequence $T = [t_{1},\ldots,t_{N}]\in \mathbb{R}^{N\times d_{p}}$. 
A nearest-neighbor look-up is used to quantise each token $t_{i}$ into token $z_{i}$ in a codebook $\mathbf{C}$. 
Subsequently, an MLP-Mixer~\cite{tolstikhin2021mlpmixer} based \textit{decoder} is employed to decode the quantised token sequence $Z$ back to 2D poses $\hat{P} \in \mathbb{R}^{J \times 2}$. 
Exponential moving average (EMA) is used to update the codebook due to the non-differentiable problem. 
In stage two, a \textit{classifier} is trained to take a human image $X \in \mathbb{R}^{C\times H \times W}$ as input and output $\hat{L} \in \mathbb{R}^{N\times V}$ representing the likelihood of each token belonging to each entry $z_i$ in the codebook $\mathbf{C}$. 
To explicitly model the uncertainty of 2D pose, we further extract the biggest logits from likelihood $\hat{L}$ as pose confidence $S_{P} \in \mathbb{R}^{N\times1}$. 
Ground truth label $L \in \mathbb{R}^{N}$ can be obtained through the \textit{pose encoder} trained in stage one and hence cross-entropy loss can be employed. 
Pose token sequence $\hat{T}_{P} \in \mathbb{R}^{N\times d_{p}}$ is computed by multiplication between $\hat{L}$ and $\mathbf{C}$. 
Pose tokenizer is trained on in-the-wild 2D pose datasets like~\cite{lin2014coco} in order to be occlusion-robust since occlusion scenarios are universal in outdoor conditions. 
Refer to~\cite{geng2023pct} for more training and implementation details of the pose tokenizer.

We set the length of pose token sequence $N$ to 34 and the number of entries in the codebook $\mathbf{C}$ to 2048. 
It is important to emphasize that the number of tokens in the sequence does not correspond to the number of human joints. 
In our work, we adopt the publicly available\footnote{\url{https://github.com/Gengzigang/PCT}} pose tokenizer as described in~\cite{geng2023pct}; note that \textit{decoder} is not used since we focus on 3D human mesh recovery instead of 2D pose prediction.

\subsection{Overall pipeline}
The overall pipeline of \algname is depicted in ~\cref{fig:model-arch} (a). 
Our proposed method leverages two transformers to regress human mesh from single image. 
Transformer encoders are used to enable message passing between the camera token, image tokens, and pose tokens. 
The decoders extract useful information from the encoder, output joints tokens and vertex tokens, which will be decoded to human mesh. 
The encoder-decoder architecture is illustrated in ~\cref{fig:model-arch} (b). 
An extra dimensionality reduction operation is conducted between transformers to reduce computations. 

\noindent\textbf{Feature Extraction} Given a cropped human image $X \in \mathbb{R}^{C \times H \times W}$, a CNN backbone is employed to extract feature map $X_{I} \in \mathbb{R}^{d_{i} \times h \times w}$, note that $d_{i}$ represents size of the channel dimension and $h \times w$ represents the spatial size. 
Typical values we use are $d_{i} = 2048$ and $h\times w=\frac{H}{32}\times \frac{W}{32}$. 
We concatenate pose token $\hat{T}_{P}$ and pose confidence $S_{P}$. Then, we utilize 4 MLP-Mixer-based~\cite{tolstikhin2021mlpmixer} layers as the \textbf{\textit{pose feature modulator}} to adjust the channel dimension size of their combination, resulting in a modulated sequence $T_{P} \in \mathbb{R}^{N\times d}$. Detailed architecture of our MLP-Mixer-based layer is provided in the supplementary material.

\noindent\textbf{Transformer Encoder} First, a $1\times1$ convolution is used as the \textbf{\textit{image feature modulator}} to reduce the channel dimension of feature map $X_{I}$, from $d_i$ to a smaller dimension $d$. 
We flatten and transpose the reduced feature map and obtain image token sequence $T_{I} \in \mathbb{R}^{hw \times d}$. 
Transformer encoder of \algname takes image tokens along with learnable camera token $T_{C} \in \mathbb{R}^{1 \times d}$ and modulated pose tokens $T_{P} \in \mathbb{R}^{N\times d}$ as input, depicted in ~\cref{fig:model-arch} (b). 
After that, transformer encoder outputs camera feature and image-pose feature $T_{IP} \in \mathbb{R}^{(hw+N)\times d}$. 
The obtained image-pose feature will be used as key and value in transformer decoder, while the camera feature will be solely decoded by a linear layer (not shown in ~\cref{fig:model-arch}), resulting in a scaling factor $s \in \mathbb{R}$ and a translation vector $t \in \mathbb{R}^{2}$, which can be used for loss computation.

\noindent\textbf{Transformer Decoder} Transformer decoder of \algname takes both joint tokens $\mathrm{T}_{J}\in \mathbb{R}^{K\times d}$ and vertex tokens $\mathrm{T}_{V}\in \mathbb{R}^{Q\times d}$ as input. 
Following~\cite{cho2022FastMetro,kim2023PointHMR}, we apply an attention mask derived from the adjacency matrix of human mesh vertices in the self-attention module. 
The output joint feature and vertex feature are then projected into 3D coordinates by a linear 3D coordinates regressor. 
As shown in ~\cref{fig:model-arch} (a), a pre-defined mesh upsampler~\cite{ranjan2018meshupsample} $U \in \mathbb{R}^{M\times Q}$ is further applied to upsample the coarse mesh vertices $V_{c} \in \mathbb{R}^{Q \times 3}$ to fine mesh vertices $V_{f} \in \mathbf{R}^{M \times 3}$ , \textit{i.e.}, $V_{f} = UV_{c}$.
Typical values we use are $K = 14$, $Q = 431$ and $M = 6890$.

\subsection{Loss Design}
As a non-parametric method, the loss function we employ to train \algname incorporates three types of supervisory signals: 3D vertex loss $L_v$, 3D joints loss $L_{3D}$, and 2D joints loss $L_{2D}$. 

To penalize the difference between 3D coordinates of predicted human body mesh vertices and corresponding ground truth, $L_1$ loss is adopted and $L_v$ could be represented as follows:
\vspace{-0.45\baselineskip}
\begin{equation}
  L_v = \frac{1}{M}\sum_{i=1}^{M}\|\hat{V}^i - V^i\|_1,
  \vspace{-0.45\baselineskip}
  \label{eq:vertex_loss}
\end{equation}
where \(\hat{V}^i\) denotes the predicted \(i\)th vertex, and \(V^i\) represents the corresponding ground truth. \(M\) denotes the total number of human mesh vertices.

Following~\cite{lin2021Metro, cho2022FastMetro}, $L_{3D}$ is also used to penalize the error of human joints in 3D space. 
It is worth noting that in our experiments, predicted 3D coordinates of human joints come from two different sources: $\Bar{J}_{3D}$ which is obtained by decoding the joint token using linear layer and $\hat{J}_{3D}$ which is obtained by applying pre-computed joint regression matrix $M_{J}$ on human mesh $\hat{V}$, \textit{i.e.}, $\hat{J}_{3D} = M_{J}\hat{V}$. 
This loss can be represented as follows:

\vspace{-0.4\baselineskip}
\begin{equation}
  L_{3D} = \frac{1}{K}\sum_{i=1}^{K}(\|\hat{J}_{3D}^i - {J}_{3D}^i\|_1 + \|\Bar{J}_{3D}^i - {J}_{3D}^i\|_1),
  \vspace{-0.4\baselineskip}
  \label{eq:j3d_loss}
\end{equation}
where \({J}_{3D}\) denotes the ground truth 3D coordinates of human joints. \(K\) denotes the total number of human joints.

To further assure the alignment between 2D image clue and 3D human joints and leverage the abundant accurate 2D key-points annotation in multiple datasets~\cite{ionescu2013human3,lassner2017up3d,mehta2018muco,lin2014coco,andriluka2014mpii,von2018pw3d},~\cite{kanazawa2018hmr, kolotouros2019spin} proposed to applying 2D joints loss $L_{2D}$ to supervise model. Following them, we apply orthogonal projection operation $\Pi(\cdot)$ on 3D joints $\Bar{J}_{3D}$, $\hat{J}_{3D}$ to obtain 2D joints $\Bar{J}_{2D}$, $\hat{J}_{2D}$ coordinates by using camera parameters $\{s,t\}$ decoded from camera token as follows:
\vspace{-0.4\baselineskip}
\begin{equation}
  \hat{J}_{2D} = s\Pi(\hat{J}_{3D}) + t, \ \ \ 
  \Bar{J}_{2D} = s\Pi(\Bar{J}_{3D}) + t
  \vspace{-0.4\baselineskip}
  \label{eq:orthogonal_proj_2};
\end{equation}
hence the 2D loss can be represented as follows:
\vspace{-0.4\baselineskip}
\begin{equation}
  L_{2D} = \frac{1}{K}\sum_{i=1}^{K}(\|\hat{J}_{2D}^i - {J}_{2D}^i\|_1 + \|\Bar{J}_{2D}^i - {J}_{2D}^i\|_1).
  \vspace{-0.4\baselineskip}
  \label{eq:j2d_loss}
\end{equation}

To summarize, our overall loss can be represented as :
\vspace{-0.4\baselineskip}
\begin{equation}
  L_{total} = \alpha_{1}L_v + \alpha_{2}L_{3D} + \alpha_{3}L_{2D},
  \vspace{-0.4\baselineskip}
  \label{eq:total_loss}
\end{equation}
where $\alpha_{1}$, $\alpha_{2}$, $\alpha_{3}$ are all weight terms, which are empirically set to 1000, 1000 and 100, respectively.
\section{Experiments}
\label{sec:Experiment}
\subsection{Experimental Setup}
\label{dataset}
\textbf{Datasets.}
Following prior works~\cite{lin2021Metro,cho2022FastMetro,Cho2023diffhmr}, we adopt Human3.6M~\cite{ionescu2013human3}, UP-3D~\cite{lassner2017up3d}, MuCo-3DHP~\cite{mehta2018muco} as 3D datasets and pseudo-ground-truth SMPL annotated COCO~\cite{lin2014coco} and MPII~\cite{andriluka2014mpii} as 2D datasets for training our model. 
For fair comparison, when testing on 3DPW-OCC~\cite{zhang2020pw3docc}, 3DPW-PC~\cite{sun2021romp} and 3DPW-ALL~\cite{von2018pw3d}, we do not fine-tune our model on any split of 3DPW~\cite{von2018pw3d} dataset. 
When testing on 3DOH~\cite{zhang2020pw3docc}, we fine-tune our model on 3DOH training set following~\cite{kocabas2021pare,li2023niki}. 
When testing on 3DPW-TEST~\cite{von2018pw3d} split, we fine-tune our model with 3DPW-TRAIN~\cite{von2018pw3d} split like~\cite{lin2021Metro, cho2022FastMetro, lin2021MeshGraphormer, lin2022mpt} to comprehensively examine the performance gap of the model compared to the existing state-of-the-art methods.
For each input image, the region containing target human is cropped using annotated bounding boxes and resized to $256 \times 256$. 
We only apply color jittering, random rotation and scaling to cropped images as data augmentation.

\noindent\textbf{Implementation Details.}
All the experiments are implemented with PyTorch. 
We train our model on NVIDIA RTX3090 GPUs for 30 epochs.
The total batch size is 128 when using ResNet-50 and 96 when using HRNet-W48. 
AdamW optimizer is adopted with its momentum factors set to 0.9 and 0.999. 
The learning rate is set to $2 \times 10^{-5}$ during fine-tuning on the 3DPW-TRAIN split, and is set to $1 \times 10^{-4}$ by default otherwise.
Note that in all tables, if not otherwise stated, the results of our methods are obtained after fine-tuning on the training set.

\noindent\textbf{Evaluation Metrics.}
We adopt three commonly used metrics: mean per vertex position error (\textbf{MPVPE}), mean per joint position error (\textbf{MPJPE}) and Procrustes-aligned mean per joint position error (\textbf{PA-MPJPE}). 
\textbf{MPVPE} measures the alignment between predicted human mesh vertices and its ground truth, while \textbf{MPJPE} and \textbf{PA-MPJPE} measure the accuracy of the recovered 3D human poses.


\begin{table*}[!t]
  \centering
    \renewcommand{\tabcolsep}{4pt}
    \vspace{-4mm}
  \begin{tabular}{@{}lcccc@{}}
    \toprule
    Methods & Backbone & \multicolumn{3}{c}{3DPW-TEST} \\
    \cmidrule(lr){3-5}
    && MPVPE($\downarrow$) & MPJPE($\downarrow$) & PA-MPJPE($\downarrow$) \\
    \midrule
    HMR\textsuperscript{\textdagger}~\cite{kanazawa2018hmr} & ResNet-50 & - & 130.0 & 81.3 \\
    SPIN\textsuperscript{\textdagger}~\cite{kolotouros2019spin} & ResNet-50 & 116.4 & 96.9 & 59.2 \\
    VIBE~\cite{kocabas2020vibe} & ResNet-50 & 99.1 & 82.9 & 51.9 \\
    3DCrowdNet\textsuperscript{\textdagger}~\cite{choi2022_3DCrowdNet} & ResNet-50 & 98.3 & 81.7 & 51.5 \\ 
    ROMP~\cite{sun2021romp} & ResNet-50 & 94.7 & 79.7 & 49.7 \\
    HybrIK~\cite{li2021hybrik} & ResNet-34 & 86.5 & 74.1 & 45.0 \\
    VisDB~\cite{yao2022visdb} & ResNet-50 & 83.5 & 72.1 & 44.1 \\
    INT-2~\cite{yang2023capturing} & ResNet-50 & 87.9 & 75.6 & 42.0 \\
    \midrule
    Ours(w/o pose tokens) & ResNet-50 & 82.5 & 72.4 & 44.4 \\
    Ours & ResNet-50 & \textbf{78.0}\textcolor{blue}{($\downarrow$5.5\%)} & \textbf{68.4}\textcolor{blue}{($\downarrow$5.5\%)} & \textbf{40.8}\textcolor{blue}{($\downarrow$8.1\%)} \\
    \midrule
    Pose2Mesh~\cite{choi2020pose2mesh} & HRNet-W48 & 105.3 & 89.5 & 56.3 \\
    PARE~\cite{kocabas2021pare} & HRNet-W32 & 88.6 &  74.5 & 46.5 \\
    METRO~\cite{lin2021Metro} & HRNet-W64 & 88.2 & 77.1 & 47.9\\
    MeshGraphormer~\cite{lin2021MeshGraphormer} & HRNet-W64 & 87.7 & 74.7 & 45.6\\
    FastMETRO~\cite{cho2022FastMetro} & HRNet-W64 & 84.1 & 72.5 & 44.6 \\
    PSVT~\cite{qiu2023psvt} & HRNet-W32 & 84.0 & 73.1 & 43.5 \\ 
    CLIFF~\cite{li2022cliff} & HRNet-W48 & 81.2 & 69.0 & 43.0 \\
    MPT~\cite{lin2022mpt} & HigherHRNet & 79.4 & \textbf{65.9} & 42.8 \\
    VirtualMarker~\cite{Ma2023virtual} & HRNet-W48 & 77.9 & 67.5 & 41.3 \\
    NIKI~\cite{li2023niki} & HRNet-W48 & 81.6 & 71.3 & 40.6 \\
    \midrule
    Ours(w/o pose tokens) & HRNet-W48 & 78.6 & 68.9 & 41.5 \\
    Ours & HRNet-W48 & \textbf{76.8}\textcolor{blue}{($\downarrow$2.3\%)} & 67.7\textcolor{blue}{($\downarrow$1.7\%)} & \textbf{39.8}\textcolor{blue}{($\downarrow$4.1\%)} \\
    \bottomrule
  \end{tabular}
  \vspace{-2mm}
  \caption{Performance on 3DPW~\cite{von2018pw3d}. Backbones are denoted for fair comparison. Results in bold indicate the best performance. Note that \textsuperscript{\textdagger} denotes the performance \textbf{\textit{without}} fine-tuning on the 3DPW train-split. Text in \textcolor{blue}{blue} indicates the performance improvement compared with w/o pose tokens baseline. }
    \vspace{-2mm}
  \label{tab:pw3d}
\end{table*}

\begin{table*}[!t]
  \centering
    \renewcommand{\tabcolsep}{3pt}
  \begin{tabular}{@{}lccccc@{}}
    \toprule
    Methods & \multicolumn{3}{c}{3DPW-OCC} & \multicolumn{2}{c}{3DOH}\\
    \cmidrule(lr){2-4} \cmidrule(lr){5-6}
    & MPVPE($\downarrow$) & MPJPE($\downarrow$) & PA-MPJPE($\downarrow$) & MPJPE($\downarrow$) & PA-MPJPE($\downarrow$)  \\
    \midrule
    Zhang \textit{et al.}~\cite{zhang2020pw3docc} & - & - & 72.2 & - & 58.5 \\
    SPIN~\cite{kolotouros2019spin} & 121.6 & 95.6 & 60.8 & 104.3 & 68.3 \\
    HybrIK~\cite{li2021hybrik} & 111.9 & 90.8 & 58.8 & 40.4 & 31.2 \\
    HMR-EFT~\cite{joo2021exemplar} & 111.3 & 94.4 & 60.9 & 75.2 & 53.1 \\
    PARE~\cite{kocabas2021pare} & 107.9 & 90.5 & 56.6 & 60.9 & 42.7 \\
    NIKI~\cite{li2023niki} & 107.6 & 85.5 & 53.5 & 38.8 & 28.7 \\
    \midrule
    Ours (ResNet-50)* & 93.5 & 84.0 & 50.0 & 39.2 & 28.7 \\
    Ours (HRNet-W48)* & \textbf{90.1} & \textbf{79.7} & \textbf{49.0} & \textbf{35.4} & \textbf{26.8} \\
    \bottomrule
  \end{tabular}
  \vspace{-2mm}
  \caption{Performance on 3DPW-OCC~\cite{zhang2020pw3docc} and 3DOH~\cite{zhang2020pw3docc}. Note that all methods are tested \textbf{\textit{without}} any fine-tuning on 3DPW dataset. Results in bold indicate the best performance. * means extra datasets used compared to PARE~\cite{kocabas2021pare}. Note that we fine-tune our model on 3DOH \textbf{only} when testing on 3DOH.}
  \vspace{-4mm}
  \label{tab:pw3d-occ}
\end{table*}

\begin{table}[!t]
  \centering
  \small
    \renewcommand{\tabcolsep}{4.5pt}
  \begin{tabular}{@{}lccc@{}}
    \toprule
    Methods & \multicolumn{3}{c}{3DPW-PC} \\
    \cmidrule(lr){2-4}
    & MPVPE($\downarrow$) & MPJPE($\downarrow$) & PA-MPJPE($\downarrow$) \\
    \midrule
    VIBE & - & - & 103.5 \\
    SPIN~\cite{kolotouros2019spin} & 157.6 & 129.6 & 82.6 \\
    PyMaf~\cite{zhang2021pymaf} & 154.3 & 126.7 & 81.3 \\
    ROMP~\cite{sun2021romp} & 147.5 & 115.6 & 75.8 \\
    OCHMR~\cite{khirodkar2022ochmr} & 145.9 & 112.2 & 75.2 \\
    \midrule
    Ours (ResNet-50) & 113.4 & 97.8 & 62.4  \\
    Ours (HRNet-W48) & \textbf{110.9} & \textbf{95.3} & \textbf{61.0}  \\
    \bottomrule
  \end{tabular}
  \caption{Performance on 3DPW-PC~\cite{sun2021romp}. Note that all methods are tested \textbf{\textit{without}} any fine-tuning on 3DPW dataset. Results in bold indicate the best performance.}
  \vspace{-6mm}
  \label{tab:pw3d-pc}
\end{table}

\begin{figure*}[t]
  \centering
   \includegraphics[width=1.0\linewidth]{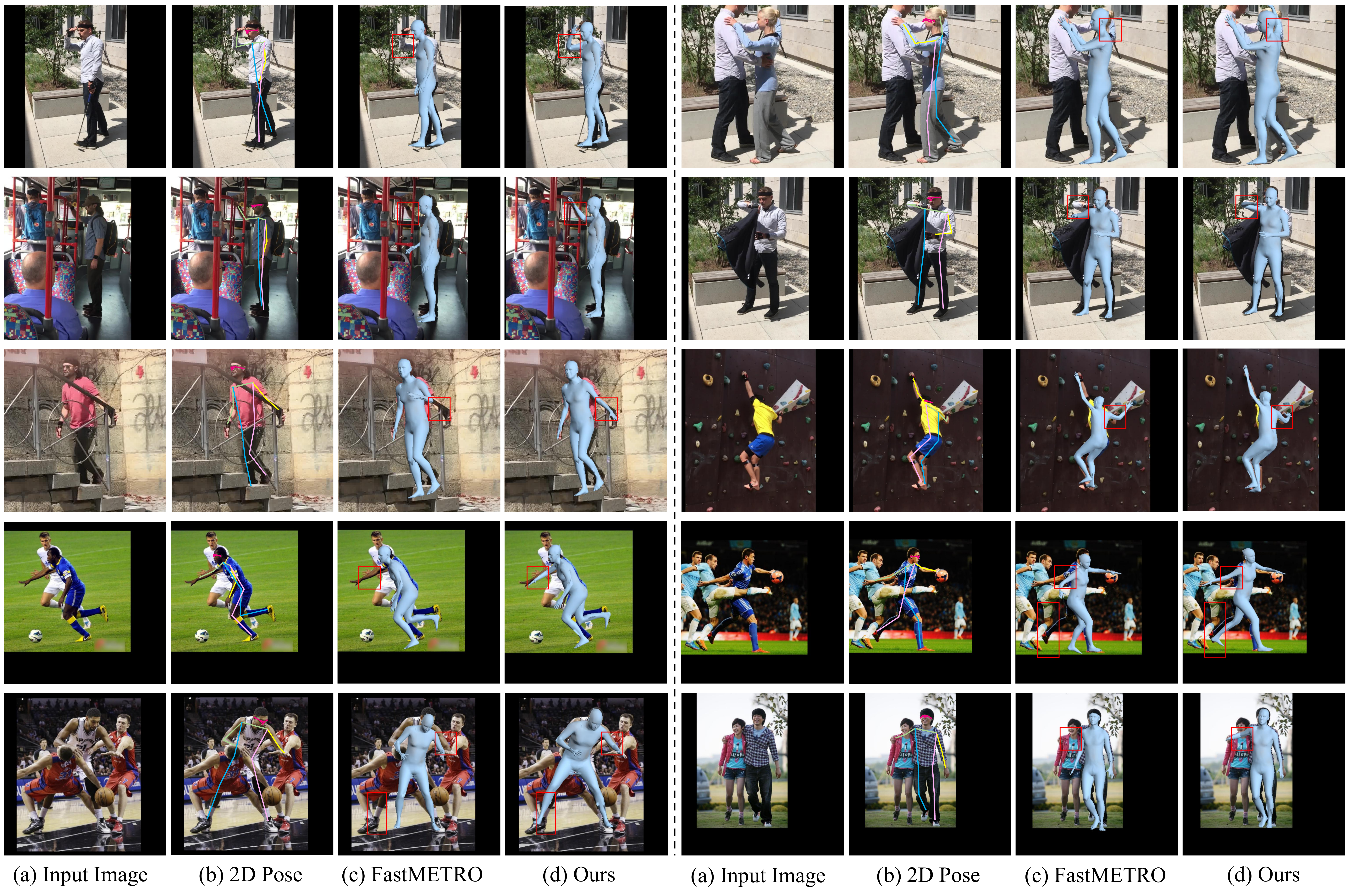}
   \caption{\textbf{Qualitative results on 3DPW (rows 1-3) and OCHuman (rows 4-5) datasets.} From left to right: (a) Input image, (b) 2D Pose decoded from pose tokens, (c) FastMETRO~\cite{cho2022FastMetro} results, (d) Our results.}
   \vspace{-2mm}
   \label{fig:postmetro-vis}
\end{figure*}

\begin{table*}[!t]
  \centering
   \renewcommand{\tabcolsep}{6pt}
    \begin{tabular}{@{}cccccc@{}}
    \toprule
    Method & Training time & Inference time & \multicolumn{3}{c}{3DPW-TEST} \\
    \cmidrule(lr){4-6}
    &&& MPVPE($\downarrow$) & MPJPE($\downarrow$) & PA-MPJPE($\downarrow$) \\
    \midrule
    METRO~\cite{lin2021Metro} & $\sim$1400 GPU hrs & 15.8FPS & 88.2 & 77.1 & 47.9 \\
    MeshGraphormer~\cite{lin2021MeshGraphormer} & $\sim$1300 GPU hrs & 15.0FPS & 87.7 & 74.7 & 45.6 \\
    FastMETRO~\cite{cho2022FastMetro} & \textbf{$\sim$330GPU hrs} & 17.0FPS & 84.1 & 72.5 & 44.6 \\
    Ours (ResNet-50) & $\sim$340 GPU hrs & \textbf{18.7FPS} & 78.0 & 68.4 & 40.8 \\
    Ours (HRNet-W48) & $\sim$390 GPU hrs & 11.5FPS & \textbf{76.8} & \textbf{67.7} & \textbf{39.8} \\
    \bottomrule
  \end{tabular}
  \caption{Training and Inference time compared with other baselines. Our proposed method shows competitive efficiency while improving performance.}
  \label{tab:training-infer-time}
\end{table*}

\begin{table*}[!t]
  \centering
   \renewcommand{\tabcolsep}{6pt}
    \begin{tabular}{@{}cccccccc@{}}
    \toprule
    Image & Pose & \multicolumn{2}{c}{3DPW-OCC} & \multicolumn{2}{c}{3DPW-PC} & \multicolumn{2}{c}{3DPW-ALL} \\
    \cmidrule{3-4} \cmidrule(lr){5-6} \cmidrule(lr){7-8}
     Tokens & Tokens & MPJPE($\downarrow$) & PA-MPJPE($\downarrow$) & MPJPE($\downarrow$) & PA-MPJPE($\downarrow$) & MPJPE($\downarrow$) & PA-MPJPE($\downarrow$)\\
    \midrule
     & \ding{51} & 90.2 & 56.4 & 104.1 & 65.7 & 93.3 & 54.5 \\
    \ding{51} &  & 92.1 & 56.3 & 111.0 & 71.0 & 91.9 & 53.2 \\
    \ding{51} & \ding{51} & \textbf{84.0} & \textbf{50.0} & \textbf{97.8} & \textbf{62.4} & \textbf{83.1} & \textbf{48.2}  \\
    \bottomrule
  \end{tabular}
  \caption{Quantitative comparisons on 3DPW-OCC~\cite{zhang2020pw3docc}, 3DPW-PC~\cite{sun2021romp} and 3DPW-ALL~\cite{von2018pw3d}. Note that we use ResNet-50 as our backbone. Results are obtained \textbf{\textit{without}} fine-tuning on 3DPW-TRAIN split and those in bold indicate the best performance.}
  \vspace{-2mm}
  \label{tab:pw3d-ablation}
\end{table*}

\subsection{Main Results}
\textbf{Quantitative Results.} To validate the effectiveness of our model under different settings, we test our model on different datasets, \textit{i.e.}, 3DPW-TEST, 3DPW-OCC~\cite{zhang2020pw3docc}, 3DOH~\cite{zhang2020pw3docc} and 3DPW-PC~\cite{sun2021romp}. 

We first report the performance of our model compared to other baselines on the 3DPW-TEST split in~\cref{tab:pw3d}. 
We fine-tune our model on 3DPW-TRAIN split following prior works~\cite{lin2021Metro,cho2022FastMetro,lin2021MeshGraphormer,kim2023PointHMR}. 
As can be seen, our model with an HRNet-W48 backbone achieves SOTA performance on 3DPW-TEST dataset, with 76.8mm MPVPE, 67.7mm MPJPE and 39.8mm PA-MPJPE. 
Despite that our proposed method is indeed inferior to MPT~\cite{lin2022mpt} and VirtualMarker~\cite{Ma2023virtual} on MPJPE, it still beats all SOTA methods by a noticeable margin on each MPVPE and PA-MPJPE, showing the capability of reconstructing human mesh on the in-the-wild dataset. 
It's worth noting that MPT benefits from a much larger pre-training dataset, which could explain the performance gap. For more details on dataset scale comparison, please see the supplementary materials.
It is also worth noting that even when substituting our backbone to a more light-weighted one, \textit{i.e.}, ResNet-50, compared to other methods using the same backbone, our model still demonstrates superior results. 
Specifically, our method outperforms those with special designs for extreme cases including truncation and occlusion, like VisDB~\cite{yao2022visdb} and 3DCrowdNet~\cite{choi2022_3DCrowdNet}.

\textbf{To stay fair, we would like to point out some key distinctions between our baseline model, called FastMETRO, and our variant without pose tokens.} 
Although both models share the same architecture, there's a notable difference in their results. The reasons are mainly three-fold: 
1. The resolution of images increases from $224 \times 224$ to $256 \times 256$. 
2. The datasets we used are refined with more pseudo-ground-truth SMPL annotations. 
3. We use backbones pretrained on COCO~\cite{lin2014coco} rather than ImageNet~\cite{deng2009imagenet}. 
For more details, please refer to supplementary material. 
We highlight that despite the above points bringing extra improvements, the effectiveness of pose tokens is still remarkable. When using ResNet-50, by simply adding pose tokens, \algname outperforms our w/o pose tokens variant by 5.5\%, 5.5\%, and 8.1\%  on MPVPE, MPJPE and PA-MPJPE, respectively. 
Despite the decrease in the extent of improvement when using HRNet-W48, comparing with such a strong baseline, achieving such enhancement remains non-trivial, particularly with a 4.1\% performance gain on PA-MPJPE.

We also conduct experiments on occlusion datasets, \textit{i.e.} 3DPW-OCC, 3DOH and 3DPW-PC. 
On 3DPW-OCC, our method with HRNet-W48 backbone shows superior performance and beats other SOTA methods, including those targeted for occlusion scenarios, such as PARE~\cite{kocabas2021pare} and NIKI~\cite{li2023niki}, by scoring 90.1mm on MPVPE, 79.7mm on MPJPE and 49.0mm on PA-MPJPE. 
On 3DOH, a similar level of superiority can also be observed. 
These consistent results on different datasets reveal the robustness of our model under the challenge of object occlusion, which further indicates the effectiveness of integrating occlusion-robust 2D pose tokens into transformer architecture.
On 3DPW-PC, our proposed methods still show competitive results, outperforming other SOTA methods including those proposed to handle person-occlusion like OCHMR~\cite{khirodkar2022ochmr} by a large margin. 
Interestingly, despite that we do not manually design any intricate components for different occlusion scenarios (\textit{i.e.}, object-occlusion and person-occlusion), our method still outperforms those hand-crafted ones~\cite{kocabas2021pare,yao2022visdb,li2023niki,khirodkar2022ochmr} and can handle different types of occlusion equally well.

\noindent\textbf{Training/Inference Time.} 
We compare \algname with METRO~\cite{lin2021Metro}, MeshGraphormer~\cite{lin2021MeshGraphormer} and FastMETRO~\cite{cho2022FastMetro}. 
We set the training parameters (e.g., epochs) as described in the original works\cite{lin2021Metro,lin2021MeshGraphormer, cho2022FastMetro}. 
The training (including pretraining) and inference times are shown in Table~\ref{tab:training-infer-time}. \algname shows competitive efficiency while improving performance.

\noindent\textbf{Qualitative Results.} We qualitatively evaluate FastMETRO~\cite{cho2022FastMetro} and our \algname on 3DPW~\cite{von2018pw3d} and OCHuman~\cite{zhang2019ochuman}. 
As in ~\cref{fig:postmetro-vis}, we observe that our \algname captures information about occluded body parts, such as limbs, as highlighted in the red rectangles. 
Compared to FastMETRO, \algname handles challenging cases in various scenarios by combining the knowledge of accurate 2D pose into 3D mesh transformer in a token-wise manner. 
This highlights the significant advantage of leveraging the spatial relationships encapsulated within 2D pose tokens for enhanced 3D mesh reconstruction.
\subsection{Ablation Studies}
\noindent\textbf{Effect of Different Tokens.}
To analyze the influence of pose tokens, we train our model with different types of tokens, \textit{i.e.}, image tokens and pose tokens, and evaluate each variant on 3DPW-OCC, 3DPW-PC and 3DPW-ALL. 
Results in ~\cref{tab:pw3d-ablation} show that, despite the naive fusion strategy, our proposed pose token enhanced model outperforms baselines by a large margin on each split.
On the 3DPW-OCC dataset, using only image or pose tokens yields inferior results. 
However, combining both as input for the transformer encoder significantly improves performance, lowering MPJPE and PA-MPJPE to 84.0mm and 50.0mm, respectively. 
These results prove that unifying these two elements as input can strengthen the robustness to occlusion and benefit 3D human mesh recovery task. 
On the 3DPW-PC dataset, using only pose tokens outperforms models that rely solely on image tokens. 
Despite not being tailored for person-occlusion scenarios, pose tokens are more effective than 2D image clues. 
Combining pose and image tokens further improves results, reducing MPJPE and PA-MPJPE to 97.8mm and 62.4mm. 
This consistent enhancement across various occlusion datasets highlights the general applicability of our method over manually designed, occlusion-specific approaches~\cite{choi2022_3DCrowdNet,kocabas2021pare}.
Moreover, a steady error decrement can also be seen on 3DPW-ALL dataset, suggesting that incorporating pose tokens and image tokens as input leads to consistent and remarkable improvement across all three dataset splits, which indicates that pose token plays a vital role in enhancing the model's ability on 3D human mesh recovery task.
Additionally, we test how well \algname performs in non-occluded scenarios. 
Our method still shows superior performance even in these cases. 
We present experimental results in non-occlusion scenarios in the supplementary materials.

\begin{table}[!t]
  \centering
  \small
      \renewcommand{\tabcolsep}{3pt}
  \begin{tabular}{@{}ccccc@{}}
    \toprule
    Backbone & Use GT & \multicolumn{3}{c}{3DPW-TEST} \\
    \cmidrule(lr){3-5}
    && MPVPE($\downarrow$) & MPJPE($\downarrow$) & PA-MPJPE($\downarrow$) \\
    \midrule
    ResNet-50 & \ding{55} & 78.0 & 68.4 & 40.8  \\
    ResNet-50 & \ding{51} & \textbf{65.9} & \textbf{57.7} & \textbf{31.3} \\
    \bottomrule
  \end{tabular}
  \caption{Performance on 3DPW~\cite{von2018pw3d} when using or not using ground truth 2D pose tokens. Results in bold indicate the best performance.}
  \label{tab:GT-ablation}
\end{table}

\noindent\textbf{Accuracy of Pose Tokens.}
For the purpose of exploring the upper limits of the token-wise 2D pose's assistance in the 3DHPSE task, we feed the tokens generated from the ground truth 2D pose by the pose encoder into the transformer. 
When using a classifier to predict pose tokens, pose confidence is generated, but it does not exist when using ground truth pose since pose encoder does not output confidence. 
Therefore, we use a very high score (fixed at 10 in the experiments) and concatenate it with the ground truth pose tokens for fine-tuning. 
The experimental results in~\cref{tab:GT-ablation} demonstrate a significant improvement in performance when utilizing ground truth pose tokens. 
When setting ResNet-50 as backbone, by using ground-truth 2D pose tokens, our method scores 65.9mm, 57.7mm and 31.3mm on MPVPE, MPJPE, PA-MPJPE respectively, highlighting the substantial benefit of accurate 2D pose in the process of 3D human mesh recovery, further implying the strong potential of our proposed methods.

\begin{table}
\centering
\begin{tabular}{ccccc}
\toprule
  Image tokens &  Pose tokens & \multicolumn{2}{c}{3DPW-TEST} \\
\cmidrule(lr){3-4}
&& MPJPE($\downarrow$) & PA-MPJPE($\downarrow$) \\
\midrule
\textit{linear} & \textit{linear} & 87.0 & 51.0   \\
\textit{linear} & \textit{mixer} & \textbf{84.9} & \textbf{48.9} \\
\textit{mixer} & \textit{mixer} & 85.4 & 50.0 \\
\bottomrule
\end{tabular}
\caption{Ablation of the modulator selection. Results are obtained \textbf{\textit{without}} fine-tuning on 3DPW-TRAIN split.}
\label{tab:modulator-sel}
\end{table}
  
\begin{table}
\centering
\renewcommand{\tabcolsep}{10pt}
\begin{tabular}{cc}
  \toprule
  Num. Blocks & 3DPW-TEST \\
  \cmidrule(lr){2-2}
  & PA-MPJPE($\downarrow$) \\
  \midrule
  1 & 50.1 \\
  2 & 49.6 \\
  4 & \textbf{48.9} \\
  8 & 49.6 \\
  \bottomrule
\end{tabular}
\captionof{table}{Ablation of mixer block number in pose feature modulator. Results are obtained \textbf{\textit{without}} fine-tuning on 3DPW-TRAIN split.}
\vspace{-5mm}
\label{tab:block-num}
\end{table}

\noindent\textbf{Ablation of Mixer layers.}
We ablate the usage of MLP-Mixer, including where to use it and how deep it should be. 
As shown in ~\cref{tab:modulator-sel}, the best results are obtained when MLP-Mixer(\textit{mixer}) is only adapted for pose tokens, while only a simple $1\times1$ convolution layer(\textit{linear}) is used for image tokens. 
In addition to performing standard matrix multiplication across channel dimensions, the MLP-Mixer uses shared weights for all channels to process data on the spatial dimension. 
This approach enhances the integration of pose tokens by improving the mixing of information among human joints. 
However, given the variability in the positioning of human body parts within images, employing fixed, shared weights for spatial operations can degrade the quality of the image features. This explains why switching from a linear modulator to a mixer modulator results in inferior performance on image features.
Results in ~\cref{tab:block-num} show that increasing the number of mixer blocks decreases the error, with the optimal performance achieved at 4 blocks. 
However, further increments deteriorate model performance, possibly due to optimization challenges.

\begin{figure}[t]
  \centering
   \includegraphics[width=1.0\linewidth]{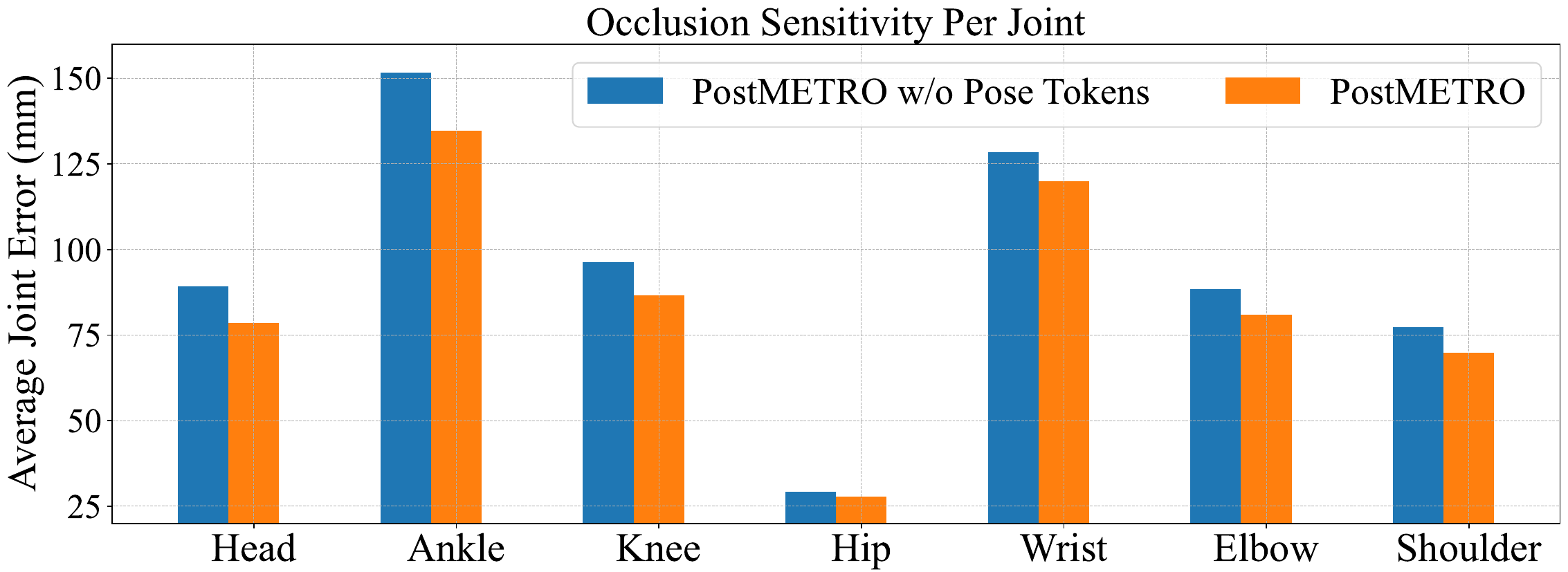}
   \caption{\textbf{Per joint occlusion sensitivity analysis} in \algname with or without joint tokens.}
   \vspace{-2mm}
   \label{fig:occlusion_sensitivity_per_joint}
\end{figure}

\noindent\textbf{Occlusion Sensitivity.}
We further follow~\cite{kocabas2021pare, li2023niki} to conduct the occlusion sensitivity analysis. 
~\cref{fig:occlusion_sensitivity_per_joint} shows the per-joint breakdown of the mean 3D error from the occlusion sensitivity analysis for \algname with or without pose tokens on the 3DPW test split. 
Combining image tokens and pose tokens, our proposed method improves robustness to occlusion in terms of all joints. 
More occlusion analyses are provided in the supplementary material.
\section{Conclusion}
\label{sec:Conclusion}
In this paper, we introduce \algname, a novel framework incorporating 2D poses and images for 3D human mesh reconstruction. 
Unlike previous methods, \algname compresses occlusion-robust 2D poses to the token level and accomplishes 3D human mesh reconstruction task by integrating pose tokens with image tokens as its input. 
This approach embeds prior knowledge of 2D poses in a data-driven manner, without relying on any manual design. 
Extensive experiments demonstrate that \algname achieves significant improvements across various datasets, including scenarios with object-occlusion, person-occlusion, and non-occlusion, which often require different design approaches. 
Moreover, our ablation studies highlight the superiority of using multiple kinds of tokens to reconstruct human mesh, and show significant performance gains when using ground truth 2D pose tokens, indicating the great potential of \algname.

{
    \small
    \bibliographystyle{ieeenat_fullname}
    \bibliography{main}
}


\clearpage
\appendix
{\Large \noindent \textbf{Appendix}} 
\bigskip

In~\cref{sec:Arch}, we first illustrate the detailed architecture of our MLP-Mixer-based layers, which are used in our pose feature modulator.
Then we provide more details about our datasets and experimental settings in~\cref{sec:datasets_supp}. 
More visualization results and occlusion sensitivity analysis results are provided in~\cref{sec:qualitative_res} and~\cref{sec:occ_analysis}. 
We also conduct more experiments in~\cref{sec:extra_exp}. 
Finally, we discuss about societal impact, our limitations and future research directions in~\cref{sec:discussion}.

\section{Architecture}
\label{sec:Arch}
\begin{figure}[!h]
  \centering
   \includegraphics[width=1\linewidth]{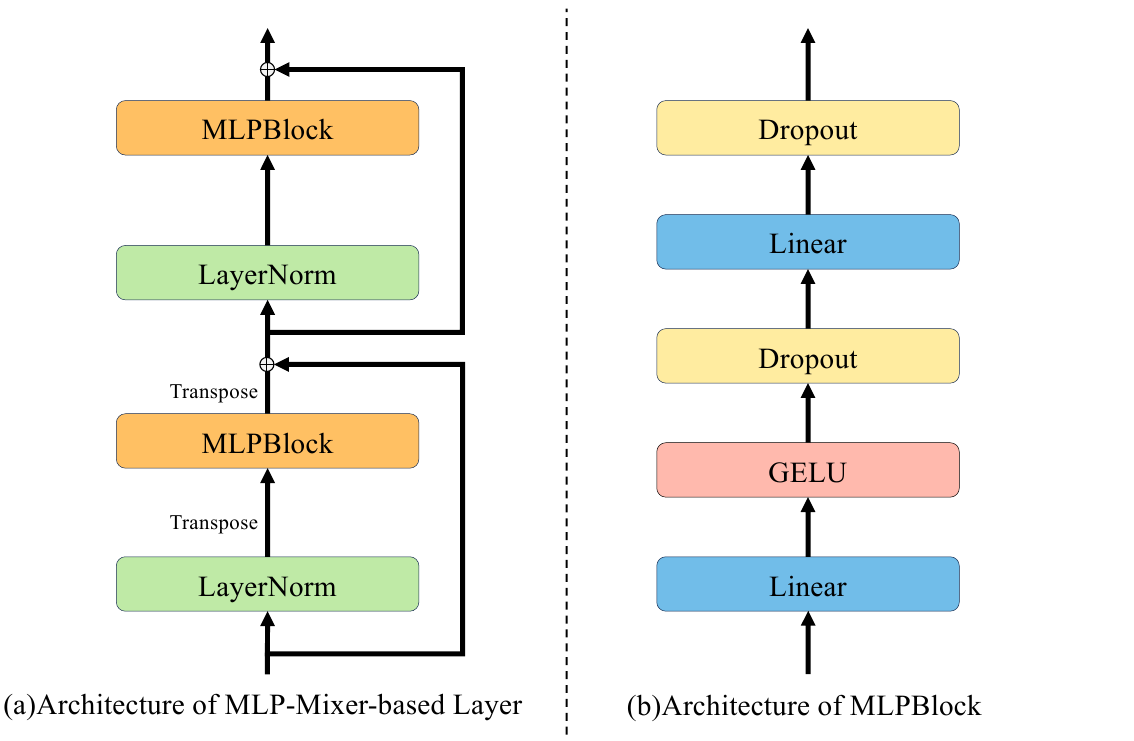}
   \caption{Detailed architecture of MLP-Mixer-based layer (a) and MLPBlock (b). }
   \vspace{-2mm}
   \label{fig:MLP-mixer-arch}
\end{figure}
The detailed architecture of our MLP-mixer-based layer is illustrated in ~\cref{fig:MLP-mixer-arch}. The MLPBlock is a key part, made up of Linear, GELU, and Dropout layers stacked together. Each MLP-Mixer-based layer contains two sets of LayerNorm and MLPBlock. Residual connection is used to ease the training.

\section{Datasets}
\label{sec:datasets_supp}
\paragraph*{3D/2D Dataset Scale}
We utilized a large mixed dataset of around 512k samples.
Those samples, including 312k from Human3.6M~\cite{ionescu2013human3}, 7k from UP-3D~\cite{lassner2017up3d}, 102k from MuCo-3DHP~\cite{mehta2018muco}. and 75k(28k + 47k) from COCO~\cite{lin2014coco}, along with 16k from MPII~\cite{andriluka2014mpii}. 
We adopt pseudo-ground-truth SMPL~\cite{SMPL:2015} annotated datasets for part of our COCO(28k) and MPII datasets from open source GitHub repository\footnote{\url{https://github.com/huawei-noah/noah-research/tree/master/CLIFF}}, and the other part of COCO(47k) from EFT\footnote{\url{https://github.com/facebookresearch/eft}} after removing items duplicated with the former 28k. 

\begin{table*}[t]
  \centering
  \small
      \renewcommand{\tabcolsep}{3pt}
  \begin{tabular}{@{}cccccc@{}}
    \toprule
    Methods & Backbone &\textit{Pre} Dataset Scale & \multicolumn{3}{c}{3DPW-TEST} \\
    \cmidrule(lr){4-6}
    &&& MPVPE($\downarrow$) & MPJPE($\downarrow$) & PA-MPJPE($\downarrow$) \\
    \midrule
    MPT & HigherHRNet & 8000K & 79.4 & \textbf{65.9} & 42.8 \\
    Ours & HRNet-W48 & 190K & \textbf{76.8} & 67.7 & \textbf{39.8} \\
    \bottomrule
  \end{tabular}
  \caption{Comparison between MPT and ours in terms of performance in the 3DPW-TEST dataset and Pretraining dataset scale. Results in bold indicate the best performance. \textit{Pre} is short for Pretraining.}
  \label{tab:mpt_comparison}
\end{table*}

\paragraph*{Pretraining Dataset Scale} 
As said in our main paper, we adopt the publicly available\footnote{\url{https://github.com/Gengzigang/PCT}} pose tokenizer as described in~\cite{geng2023pct}. 
To train this tokenizer, a large open-source codebase MMPose\footnote{\url{https://github.com/open-mmlab/mmpose}} is used and COCO 2017 dataset(150k instances) and MPII dataset(40K instances) are adopted.

In our main paper, we report the performance of MPT~\cite{lin2022mpt} and our proposed method on 3DPW-TEST dataset. 
Our proposed \algname outperforms MPT on MPVPE and PA-MPJPE metrics but underperforms on MPJPE. 
We notice that the scale of the pretraining dataset of MPT is much larger than ours, and that should be brought to concern. 
The numbers of samples of pretraining datasets are listed in ~\cref{tab:mpt_comparison}.
As can be seen, MPT leverages much more training samples, \textit{i.e.}, 80000k mesh-pose pairs, during pretraining. 
Due to such a significant gap, we believe that MPT's better performance on the MPJPE metric is partly attributed to the large-scale dataset it utilizes.

\paragraph*{Data Preprocessing} We directly adopt training data from open source GitHub repository\footnote{\url{https://github.com/microsoft/MeshTransformer}} and replace COCO and MPII data as described above. 
Besides, we download 3DPW-VAL~\cite{von2018pw3d} and 3DOH~\cite{zhang2020pw3docc} from official websites\footnote{\url{https://virtualhumans.mpi-inf.mpg.de/3DPW/}}\footnote{\url{https://www.yangangwang.com/papers/ZHANG-OOH-2020-03.html}} and then parse data. 
Note that for 3D joint annotations of 3DPW-VAL and 3DOH, we use 3D joints regressed from SMPL~\cite{SMPL:2015}. 

\paragraph*{Fine-tune Strategy} We first train \algname on mixed datasets and then fine-tune it on corresponding datasets following existing works. 
When testing on 3DPW-TEST, we fine-tune our model on 3DPW-TRAIN by setting the learning rate to $2 \times 10^{-5}$ and training it for 30 epochs. 
When testing on 3DOH, we fine-tune our model on 3DOH training split by setting the learning rate to $1 \times 10^{-4}$ and training it for 30 epochs.

When testing on 3DPW-OCC, we directly use \algname trained from mixed datasets. This differs from the policy in PARE~\cite{kocabas2021pare}, where COCO, Human3.6M, and 3DOH are used. When testing on 3DOH, we further fine-tune our model on 3DOH train set, note that we only train \algname on 3DOH train set when testing on 3DOH test set.

\section{Qualitative Results}
\label{sec:qualitative_res}
\begin{figure*}[p]
  \centering
   \includegraphics[width=1.0\linewidth]{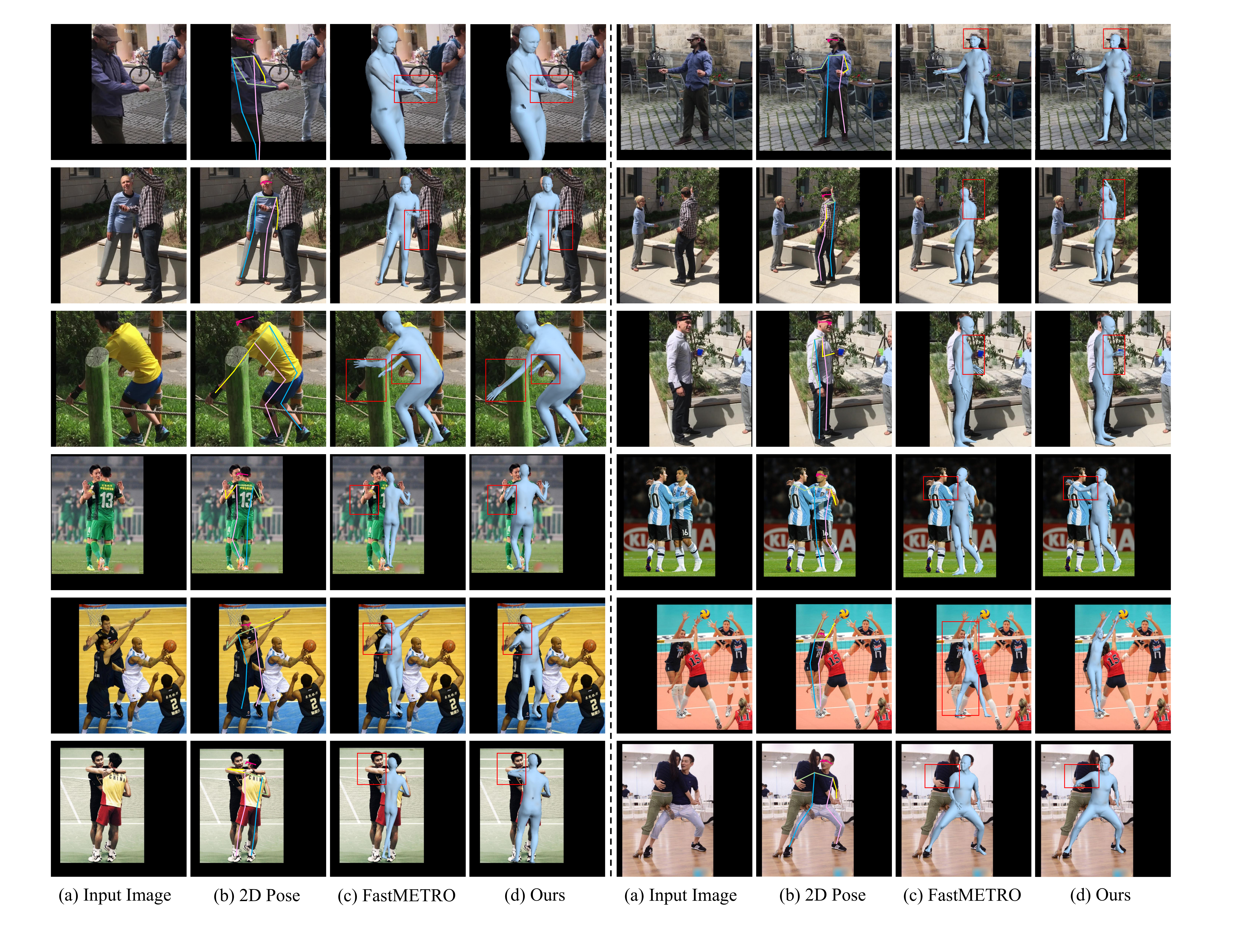}
    \vspace{-4mm}
   \caption{\textbf{Qualitative results on 3DPW (rows 1-3) and OCHuman (rows 4-6) datasets.} From left to right: (a) Input image, (b) 2D Pose decoded from pose tokens, (c) FastMETRO~\cite{cho2022FastMetro} results, (d) Our results.}
   \label{fig:extra_vis}
\end{figure*}

\begin{figure*}[t]
  \centering
   \includegraphics[width=1.0\linewidth]{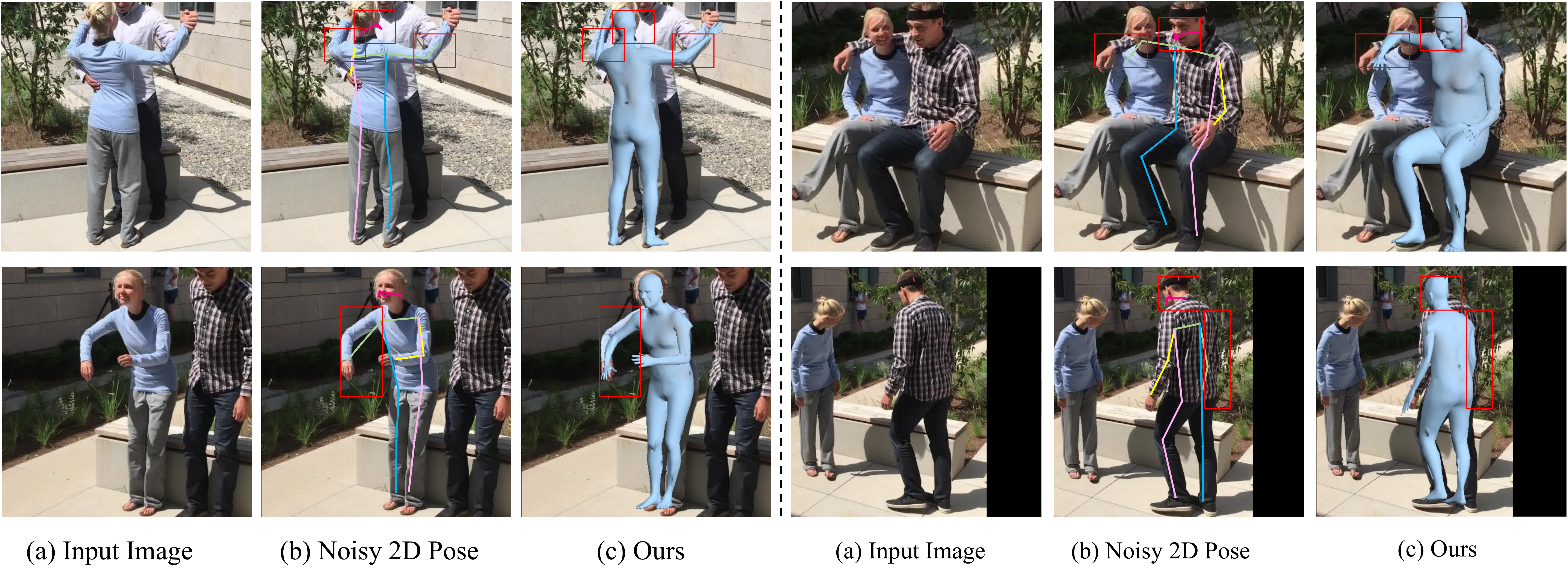}
    \vspace{-4mm}
   \caption{\textbf{Performance of \algname with noisy pose token}. From left to right: (a) Input image, (b) Noisy 2D pose, (c) Our results.}
    \vspace{-4mm}
   \label{fig:noisy_vis}
\end{figure*}

Here, we offer more qualitative results in ~\cref{fig:extra_vis}. Note that we highlight the crucial regions with red rectangular boxes. Especially, to validate that \algname does not collapse when the 2D pose is noisy, we feed noisy pose tokens to our model and test whether it can output plausible results.
First, we train our model with pose token without noise and freeze it. Then we add random Gaussian noise to the logits $\hat{L} \in \mathbb{R}^{N\times V}$ (output by the classifier in pose tokenizer) and get noisy logits $\hat{L}_{noisy} \in \mathbb{R}^{N\times V}$. We then use $\hat{L}_{noisy}$ to obtain noisy pose tokens as input for the frozen model and visualize its output. 
The output and the corresponding noisy 2D poses are shown in ~\cref{fig:noisy_vis}, we use red rectangular boxes to denote noisy regions. As can be seen, \algname can output decent results even when 2D pose is unreliable. These results demonstrate the robustness of \algname and indicate the complementary role of image tokens to pose tokens, further distinguishing our model from those that rely solely on token-wise pose representation~\cite{lin2022mpt}.

\section{Occlusion Analysis}
\label{sec:occ_analysis}
\begin{figure*}[!h]
  \centering
   \includegraphics[width=1.0\linewidth]{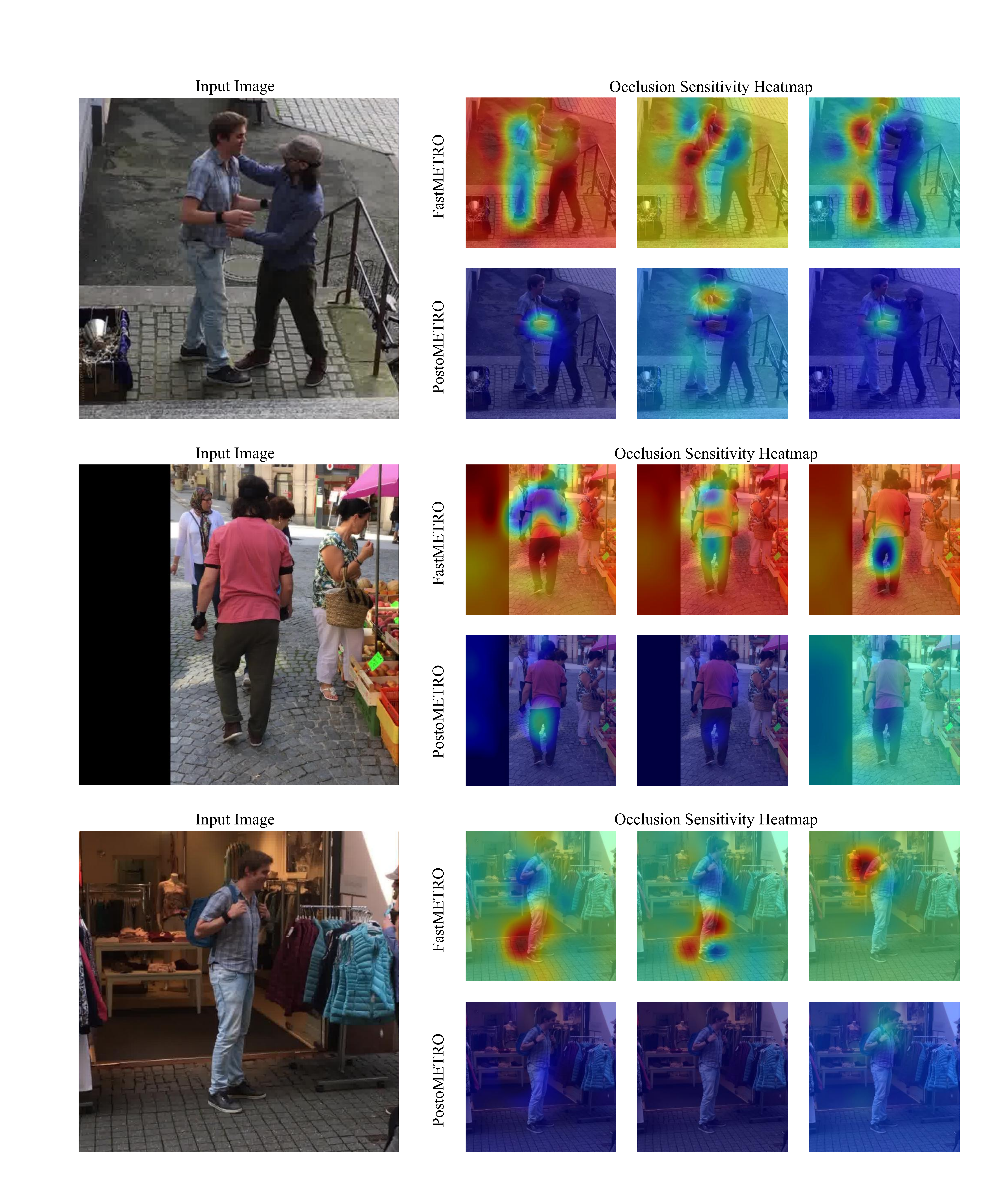}
    \vspace{-4mm}
   \caption{Occlusion Sensitivity Maps of FastMETRO~\cite{cho2022FastMetro} and \algname.}
   \label{fig:occlusion_sense_vis}
\end{figure*}

\begin{figure*}[!h]
  \centering
   \includegraphics[width=1.0\linewidth]{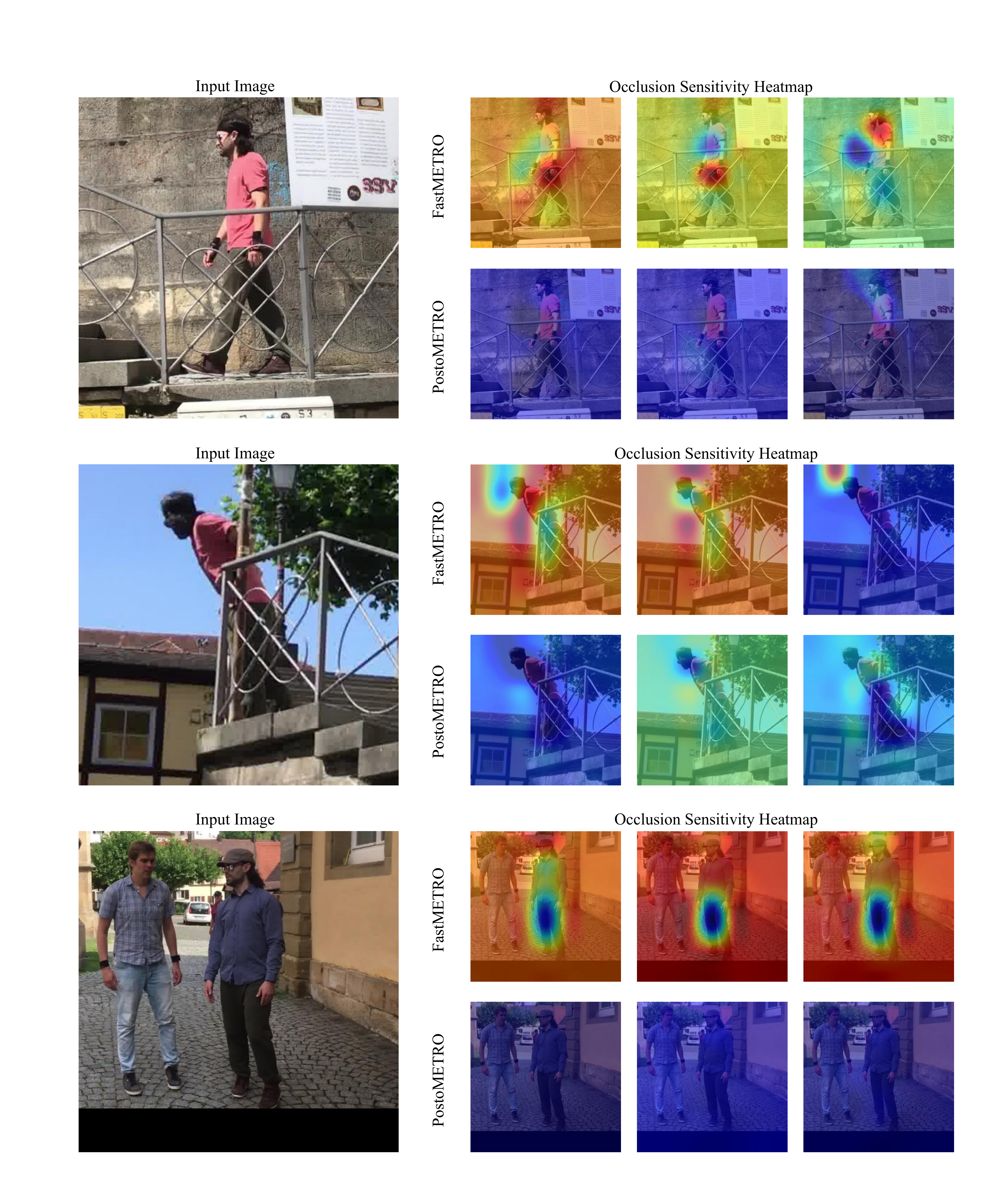}
    \vspace{-4mm}
   \caption{Additional results for Occlusion Sensitivity Maps of FastMETRO~\cite{cho2022FastMetro} and \algname.}
   \label{fig:occlusion_sense_vis_more}
\end{figure*}

Following prior works~\cite{zeiler2014visualizing,kocabas2021pare,li2023niki}, we visualize joint error maps by replacing the classification score with errors between predicted joints and corresponding ground truth. 
Same as~\cite{kocabas2021pare,li2023niki}, we use MPJPE as our measurement since PA-MPJPE leads to an artificially low error by aligning global orientations. 
We conduct our experiment on the SOTA non-parametric method, FastMETRO~\cite{cho2022FastMetro}, and our proposed \algname. 
We provide extensive results in ~\cref{fig:occlusion_sense_vis} and ~\cref{fig:occlusion_sense_vis_more}, where a warmer color denotes a higher error. 
It can be seen that \algname can produce results with lower errors in various scenarios, showing that \algname is more robust to occlusions, demonstrating its superiority.

\section{Extra ablations}
\begin{table}[!t]
  \centering
  \small
      \renewcommand{\tabcolsep}{3pt}
  \begin{tabular}{@{}cccc@{}}
    \toprule
    Methods & \multicolumn{3}{c}{3DPW-Non-OC} \\
    \cmidrule(lr){2-4}
    & MPVPE($\downarrow$) & MPJPE($\downarrow$) & PA-MPJPE($\downarrow$) \\
    \midrule
    FastMETRO~\cite{cho2022FastMetro} & 99.9 & 91.1 & 51.0 \\
    Ours & \textbf{90.1} & \textbf{82.3} & \textbf{46.8}  \\
    \bottomrule
  \end{tabular}
  \caption{Performance on 3DPW-Non-OC. Results in bold indicate the best performance. ResNet-50 is used as backbone. Note that results are obtained \textbf{\textit{without}} fine-tuning on 3DPW-TRAIN split.}
  \label{tab:nooc_results}
\end{table}

\begin{table}[!t]
  \centering
  \small
      \renewcommand{\tabcolsep}{3pt}
  \begin{tabular}{@{}cccc@{}}
    \toprule
    Pose tokens & \multicolumn{3}{c}{3DPW-TEST} \\
    \cmidrule(lr){2-4}
    & MPVPE($\downarrow$) & MPJPE($\downarrow$) & PA-MPJPE($\downarrow$) \\
    \midrule
    2D coordinates & 97.6 & 87.6 & 50.0 \\
    Ours & \textbf{91.6} & \textbf{84.9} & \textbf{48.9}  \\
    \bottomrule
  \end{tabular}
  \caption{Performance on 3DPW-TEST~\cite{von2018pw3d} when using different representations as the source of pose tokens. ResNet-50 is used as backbone. Results in bold indicate the best performance. Note that results are obtained \textbf{\textit{without}} fine-tuning on 3DPW-TRAIN split.}
  \label{tab:different_pose_tokens}
\end{table}

\label{sec:extra_exp}
\noindent\textbf{Results in non-occlusion scenarios. } We also test our model's performance under non-occlusion scenarios. 
We construct a non-occlusion subset from 3DPW by removing samples used in 3DPW-OCC and 3DPW-PC. 
The comparison between FastMETRO and PostoMETRO is listed in~\cref{tab:nooc_results}. 
Our method performs significantly better, hence proving its effectiveness even in non-occlusion scenarios.

\noindent\textbf{Results of different pose tokens.} We also try to use other type of 2D pose representation,\textit{i.e.}, raw 2D pose coordinates, as the source of pose tokens. 
We use 2D coordinates predicted by PCT~\cite{geng2023pct} and utilize the same MLP-Mixer-based layers as the pose feature modulator, with the only modification being the adjustment of input dimensions.
We train this 2D-coordinates-based-variant using the same datasets for an equal duration of training time(including pretraining).
The results are shown in ~\cref{tab:different_pose_tokens}.
As can be seen,our method, which extracts tokens from pretrained codebook as 2D pose representation, has better results.
These results prove that pretraining is necessary since it can explicitly makes structural human pose knowledge stored in the codebook, while directly end-to-end training might not guarantee such effective encoding.

\section{Discussion}
\label{sec:discussion}
\begin{figure*}[p]
  \centering
   \includegraphics[width=.8\linewidth]{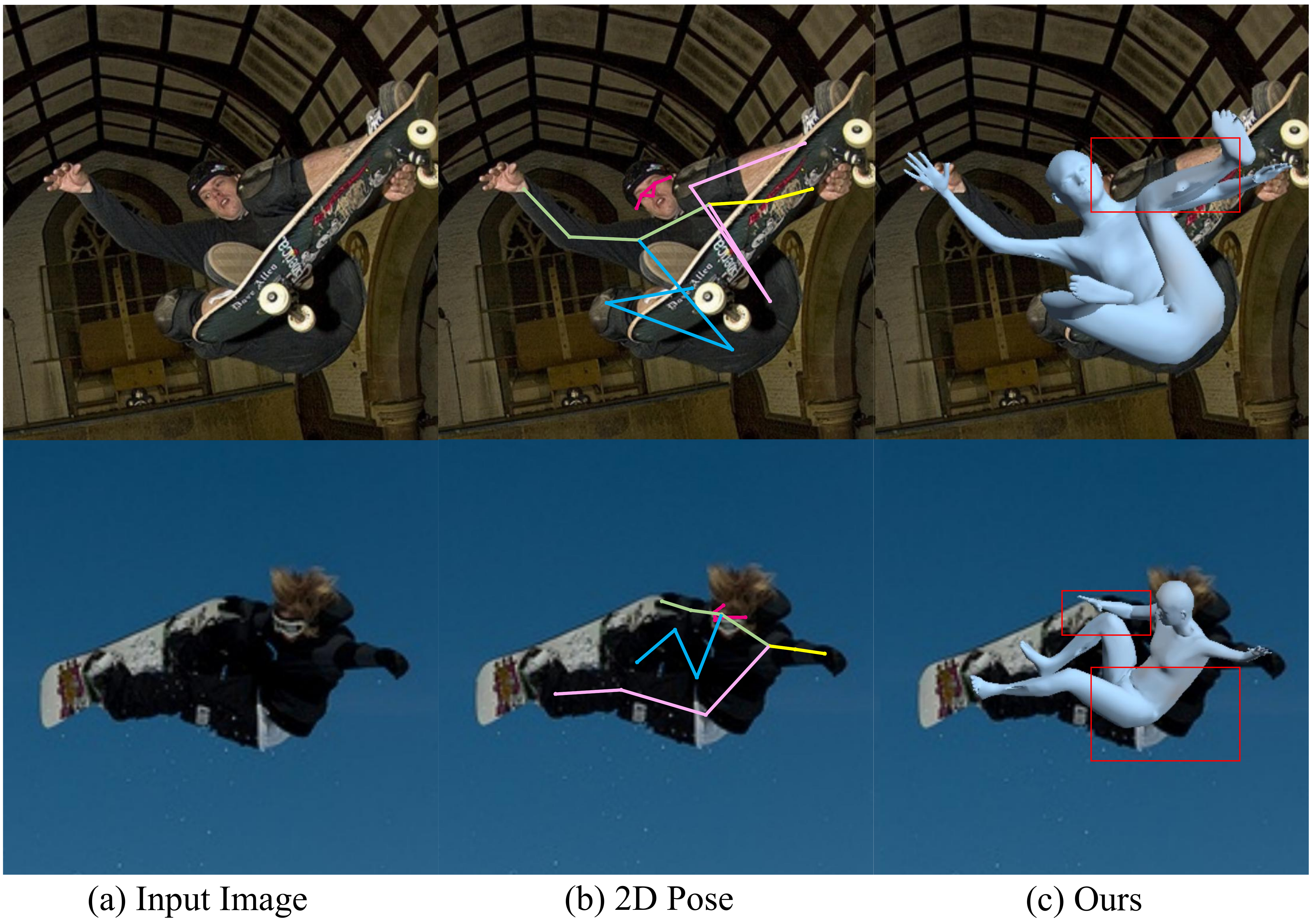}
   \caption{\textbf{Qualitative results of \algname in some challenging cases}. From left to right: (a) Input image, (b) 2D pose, (c) Our results.}
   \label{fig:limit_vis}
\end{figure*}

\noindent{\textbf{Societal Impact.}} Our proposed method can be used to detect 3D human body poses, thus applicable in certain scenarios, such as monitoring worker activities in industrial manufacturing environments or positioning patient poses in medical settings. However, in these low fault-tolerant environments, additional model assistance may be necessary when using the model.

\noindent{\textbf{Limitations \& Future research.}} Given that \algname is a data-driven approach, it may fail when there is a significant difference between the test samples and those in our datasets. 
Here we show some failure cases in~\cref{fig:limit_vis}. 
As can be seen, when the persons in the image exhibit extreme poses, \textit{e.g.}, skateboarding, \algname might not perform well and yield unsatisfactory results(\textit{e.g.}, body part misalignment bounded by red boxes), due to the lack of abundant training samples of such poses in our training sets. 
A straightforward solution is to use datasets with more diverse human poses. 
Setting that aside, exploring how to faithfully reconstruct the human mesh with extreme poses within the constraints of existing data is an interesting future work.


%
%


\end{document}